\documentclass[10pt,twocolumn,letterpaper]{article}

\usepackage{cvpr}              

\usepackage{graphicx}
\usepackage{amsmath}
\usepackage{amssymb}
\usepackage{booktabs}
\usepackage{placeins} 
\usepackage{float} 
\usepackage{enumitem}
\usepackage{tabularx} 
\usepackage{longtable} 
\usepackage{array} 
\usepackage{ragged2e} 
\usepackage{booktabs} 
\usepackage{placeins}
\usepackage{afterpage} 
\usepackage{caption} 

\usepackage[pagebackref,breaklinks,colorlinks]{hyperref}

\usepackage[capitalize]{cleveref}
\crefname{section}{Sec.}{Secs.}
\Crefname{section}{Section}{Sections}
\Crefname{table}{Table}{Tables}
\crefname{table}{Tab.}{Tabs.}

\def\cvprPaperID{*} 
\def\confName{CVPR}
\def\confYear{2022}

\begin{document}

\title{EditIDv2: Editable ID Customization with Data-Lubricated ID Feature Integration for Text-to-Image Generation}

\author{Guandong Li\\
iFlyTek\\
\and
Zhaobin Chu\\
iFlyTek\\
}

\twocolumn[{
\renewcommand\twocolumn[1][]{#1}
\maketitle
\begin{center}
    \centering
    \vspace*{-.8cm}
    \includegraphics[width=\textwidth]{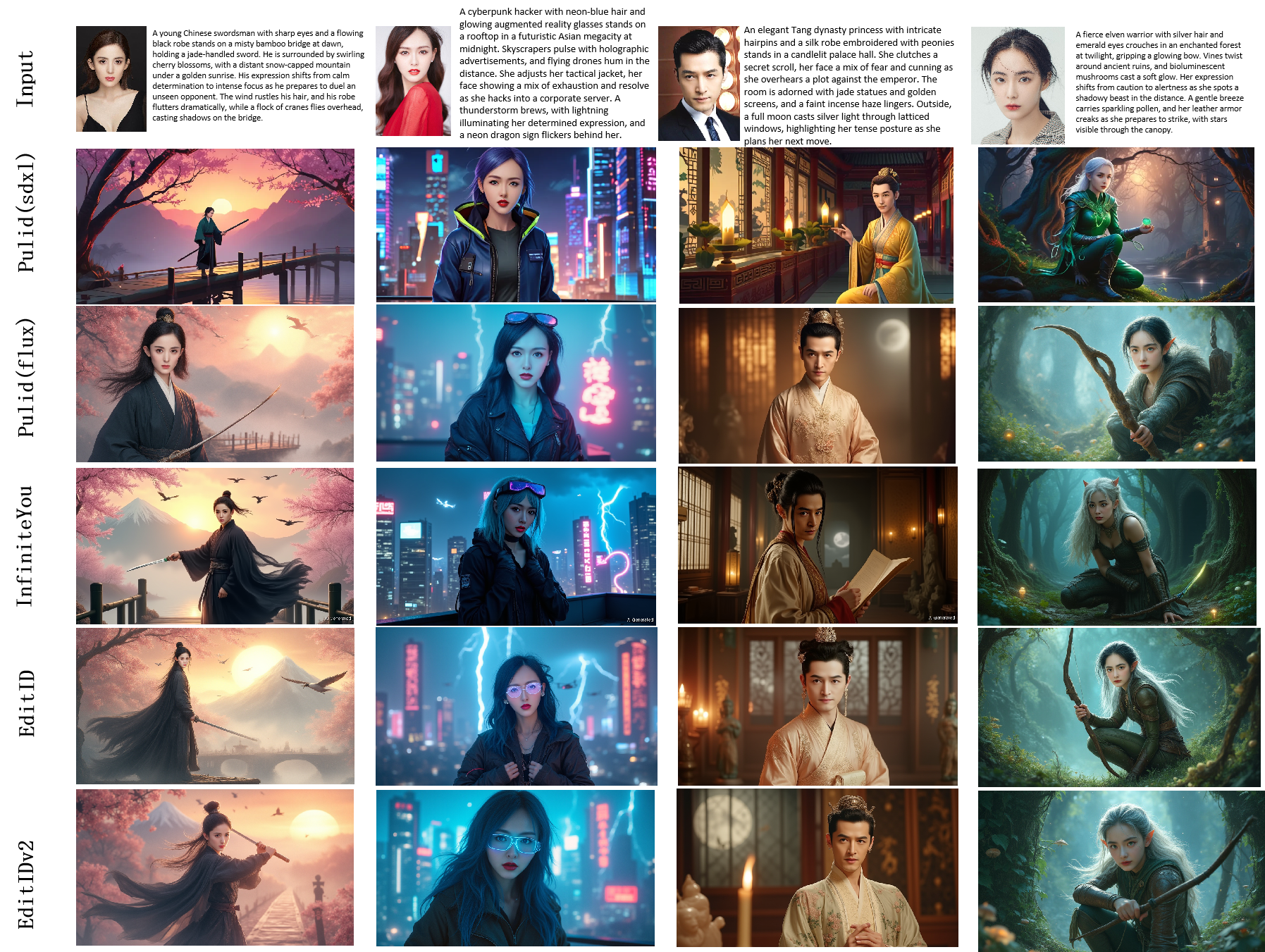}
    \vspace*{-.6cm}
    \captionof{figure}{We introduce EditIDv2, a tuning-free ID customization approach. EditIDv2 achieves better editability than similar methods in high-complexity narrative scenes and long text inputs.}
\label{fig:figure1}
\end{center}
}]

\begin{abstract}
We propose EditIDv2, a tuning-free solution specifically designed for high-complexity narrative scenes and long text inputs. Existing character editing methods perform well under simple prompts, but often suffer from degraded editing capabilities, semantic understanding biases, and identity consistency breakdowns when faced with long text narratives containing multiple semantic layers, temporal logic, and complex contextual relationships. In EditID, we analyzed the impact of the ID integration module on editability. In EditIDv2, we further explore and address the influence of the ID feature integration module. The core of EditIDv2 is to discuss the issue of editability injection under minimal data lubrication. Through a sophisticated decomposition of PerceiverAttention, the introduction of ID loss and joint dynamic training with the diffusion model, as well as an offline fusion strategy for the integration module, we achieve deep, multi-level semantic editing while maintaining identity consistency in complex narrative environments using only a small amount of data lubrication. This meets the demands of long prompts and high-quality image generation, and achieves excellent results in the IBench evaluation.
\end{abstract}

\section{Introduction}
\label{sec:intro}

Character ID customization generation \cite{li2025editid, guo2024pulid, kumari2023multi, gal2022image,ruiz2023dreambooth, liu2023facechain}, as a key branch in the field of text-to-image generation, achieves highly personalized image generation by deeply integrating specific character identity features with text prompts. It can maintain the identity features of the target character while flexibly adjusting visual attributes such as expressions, poses, and outfits in the generated image based on text descriptions, showing enormous application potential in scenarios such as digital character creation, immersive narrative construction, and personalized content production.

Many current state-of-the-art multi-subject methods \cite{tan2024ominicontrol, chen2025unireal, wu2025less, wu2025uso} attempt to use attention mechanisms in diffusion transformers (DiTs) \cite{peebles2023scalable} to inject information from reference images, but this direct injection or strong reliance on image features may significantly impact the generation quality of the base model. This often leads to artifacts, distortions, attribute entanglements, and may damage the overall structural integrity and coherence of the generated images. Current character customization methods generally focus more on character consistency, with insufficient attention to editability in high-complexity narrative scenes and long text inputs, failing to generate multi-dimensional control relative to the input character ID based on prompt changes. In EditID \cite{li2025editid}, we analyzed the reasons for the loss of editability, where the model is essentially performing an ID reconstruction task. Under a train-free \cite{tewel2024training} architecture, by deconstructing the text-to-image model for customized character IDs into an image generation branch and a character feature branch, and further decoupling the character feature branch into feature extraction, feature fusion, and feature integration modules, we achieved semantic compression of local features across network depths and formed an editable feature space by introducing a combination of mapping features and shift features, as well as controlling the input ID feature integration strength. In the feature integration module, EditID made attempts in two dimensions: first, adjusting the ID integration strength in the initial stage of denoising based on different denoising features, but this blunt adjustment ignored the progressive nature of the generation process, introducing an offset in the initial generation that damaged convergence, leading to overall darker generated images and losses in lighting and stability. Second, a softer ID strength control was chosen, considering that the editability of image generation mainly comes from the image side. However, text information is hidden in the noise image, which is essentially the starting point of the denoising process and contains latent semantic information. By compensating the noise image, additional degrees of freedom are introduced in the latent space. Reweighting is applied to the noise image query to make it the same dimension as the ID feature, followed by information supplementation through residual connection, and finally concat is chosen for fusion. Since EditID is a train-free framework, soft adjustment of feature integration is ultimately used to improve editability. EditIDv2 starts from the ID feature integration module, using a tuning-free \cite{guo2024pulid, xiao2025fastcomposer, ye2023ip} approach with a small amount of data, to more reasonably explain the method of injecting editability, significantly improving editability while maintaining character ID consistency.

The core of our method is to discuss the issue of editability injection in ID feature integration. We introduce a fine-tuning scheme for the integrated PerceiverAttention module. Assuming we already have a good image generation branch (responsible for high-quality image content generation) and character feature branch (responsible for effective extraction and processing of ID information), we focus on optimizing the information integration process from the character feature branch to the generation branch. Through a fine decomposition of the PerceiverAttention module, we break down the integration process into multiple controllable sub-stages: first, introducing an ID loss function as an auxiliary supervision signal, which ensures that identity consistency is strengthened during training by calculating the cosine similarity between the generated image and the reference ID features; second, designing a joint dynamic training mechanism for the diffusion model, dynamically adjusting the ID integration strength and depth in each step of diffusion denoising to avoid semantic conflicts caused by rigid integration in traditional train-free methods; finally, adopting an offline fusion strategy for the integration module, optimizing only the attention cross mechanism during fine-tuning, and fusing multiple optimized attention weights, which can stepwise decide whether to enhance consistency or editability, achieving efficient semantic injection without damaging the overall generalization ability of the model. Moreover, in the fine-tuning of PerceiverAttention, we found that using only a small amount of labeled data including multiple different attribute images under the same ID, editability can be gradually injected through dynamic training processes. Although the number of IDs is small, the large number of different attribute images corresponding to the ID provides rich unlocking degrees of freedom. We call this process data lubrication. Therefore, EditIDv2 solves the problem of editability injection under minimal data lubrication. Through a sophisticated decomposition of PerceiverAttention, the introduction of ID loss and joint dynamic training with the diffusion model, as well as an offline fusion strategy for the integration module, by adding a little data lubrication between the image generation and character feature branches, the capabilities of the two branches can be well utilized, and ultimately achieve SOTA results in the editability indicators of the evaluation framework IBench. In addition, although there are many methods based on the DiT architecture currently, few further optimize character editability for business scenarios such as high-complexity narrative scenes and long text inputs, so our method has great value in practical applications.

In summary, our contributions are as follows:
\begin{enumerate}
    \item Propose the EditIDv2 framework: For the tuning-free character editing problem under high-complexity narrative scenes and long text inputs, design a new editability injection scheme that significantly improves the model's semantic editing capabilities in complex contextual environments while maintaining character identity consistency.
    \item Optimize the ID feature integration module: Through a fine decomposition of the PerceiverAttention module, introduce an ID loss function, joint dynamic training of the diffusion model, and an offline fusion strategy, achieving an efficient feature integration process and solving the semantic conflict problems caused by rigid integration in traditional train-free methods.
    \item Data lubrication mechanism: Propose and verify the method of editability injection under minimal data lubrication, which can significantly improve the model's editability through dynamic training with only a small amount of labeled data, balancing efficiency and practicality.
\end{enumerate}

\section{Related Works}
\subsection{ID Consistency Generation}
With the development of T2I diffusion models, numerous personalized image generation methods have emerged. ID consistency generation, as a special branch, is mainly dedicated to capturing and restoring high-semantic-level facial features, so it has very wide applications in the field of text-to-image generation. PuLID \cite{guo2024pulid} combines a Lightning T2I branch with a standard diffusion branch, introducing contrastive alignment loss and accurate ID loss, minimizing interference with the original T2I model and ensuring good ID fidelity. EditID \cite{li2025editid} deconstructs the text-to-image model for customized character IDs, decoupling the character feature branch into feature extraction, feature fusion, and feature integration modules, achieving semantic compression of local features across network depths and forming an editable feature space by introducing a combination of mapping features and shift features, as well as controlling the input ID feature integration strength. XVerse \cite{chen2025xverse} presents a novel multi-subject control generation model. By converting reference images into offsets modulated by token-specific text flows, XVerse allows precise and independent control of specific subjects without disrupting the image latent space or features. UNO \cite{agro2024uno} consists of progressive cross-modal alignment and universal rotated position embedding. This is a multi-image conditioned subject-to-image model, iteratively trained from a text-to-image model.

\subsection{Image Editing}
Currently, few text-to-image works in character ID customization explicitly prioritize editability as a key improvement focus; more still consider the dimension of character consistency. In addition, more editing methods focus on image content editing rather than optimizing specifically for character scenes. PortraitBooth \cite{peng2024portraitbooth} leverages subject embedding features from face recognition models and incorporates emotion-aware cross-attention control to achieve diverse facial expressions in generated images, supporting text-based expression editing. DreamO \cite{mou2025dreamo} implements flexible customization of multiple image generation tasks based on pre-trained diffusion transformer (DiT) models. DreamO supports seamless integration of conditions such as identity, subject, style, and background, enhancing consistency and condition decoupling of generation results based on feature routing constraints and placeholder strategies. FLUX.1 Kontext \cite{batifol2025flux} is a novel method for image generation and editing based on flow matching \cite{lipman2022flow} technology, achieving context-aware image generation and editing in latent space. Its core goal is to generate target images through text prompts and reference images, while supporting image editing and new image generation. Current character editability methods have poor support for long prompt texts and lack flexibility.

\section{Method}
The core of our method is to solve the problem of editability injection in the ID feature integration module. Figure 2 shows the architecture of this method, deconstructing the character customization method into an image generation main branch and a character feature branch, further decoupling the character feature branch into character feature extraction, feature fusion, and feature integration modules. In EditID, we already know that ID feature integration is one of the important sources of editability, but it is not easy to control the effective and reasonable injection of editability. This section will focus on discussing how, under minimal data lubrication, through a fine decomposition of the PerceiverAttention module, the introduction of an ID loss function, joint dynamic training of the diffusion model, and an offline fusion strategy, we achieve an efficient feature integration process, solving the semantic conflict problems caused by rigid integration in traditional train-free methods.

\begin{figure*}[t]
  \centering
  \includegraphics[width=\textwidth]{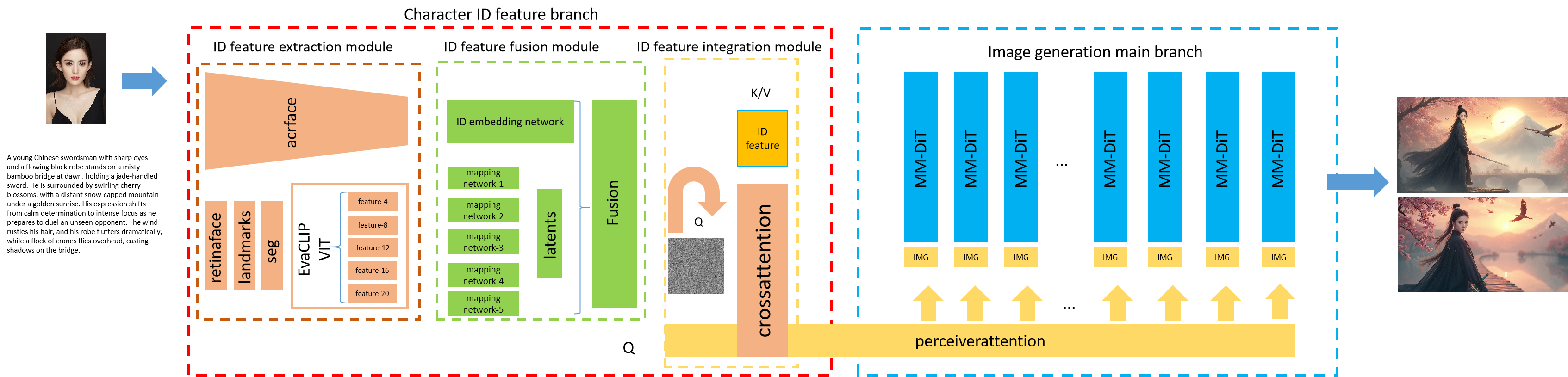}
  \caption{Overview of the EditIDv2 Framework. The left half of the framework is the character feature branch, and the right is the image generation branch. The character feature branch is divided into three parts. The first part is the ID feature extraction module, responsible for global and local feature extraction; the second part is the ID feature fusion module, responsible for mapping feature fusion; the third part is the ID feature integration module, which is the focus of the structural design in this paper.}
  \label{fig:figure2}
\end{figure*}

\subsection{Preliminary}
The essence of diffusion models is to achieve generation from simple distributions to complex data distributions by learning to reverse a process of gradually adding noise. Its forward process can be expressed as:
\begin{equation}
  x_t = a(t) \cdot x_0 + b(t) \cdot \epsilon, \quad \epsilon \sim \mathcal{N}(0, I)
\end{equation}
where $a(t)$ and $b(t)$ are schedulable functions defining the diffusion process. Based on flow matching theory \cite{esser2024scaling}, this generation process can be represented in the form of the following ordinary differential equation:
\begin{equation}
  \frac{dx}{dt} = f_\theta(x_t, t, c)
\end{equation}
Here, $f_\theta$ is the reverse vector field to be learned, and $c$ represents the conditional input (such as text embedding). Unlike traditional diffusion models, the flow matching method directly models the transformation path from noise to data, with its training objective defined as minimizing:
\begin{equation}
  \mathcal{L}(\theta) = \mathbb{E}_{t, x_0, x_1} \left[ \| f_\theta((1-t)x_0 + t x_1, t, c) - (x_1 - x_0) \|^2 \right]
\end{equation}
where $t \in [0,1]$, embodying a continuous interpolation from data ($x_0$) to noise ($x_1$).

At the architectural level, DiT (Diffusion Transformer) replaces the commonly used U-Net backbone with Transformer modules, enhancing the perception of global semantics through self-attention mechanisms. Specifically, for the conditional embedding sequence $c \in \mathbb{R}^{L \times d}$, the prediction of the conditional vector field in DiT can be expressed as:
\begin{equation}
  f_\theta = \text{Projection}(\text{MultiHeadAttention}(x_t W_Q, c W_K, c W_V))
\end{equation}
where $W_Q, W_K, W_V$ are learnable parameters. This structure can effectively capture long-distance dependencies, suitable for multi-word, long-sequence conditional generation.

This paper uses the Flux model as the baseline, which, as a generation framework based on DiT \cite{peebles2023scalable} and flow matching, proposes an improved noise scheduling strategy:
\begin{equation}
  a(t) = \cos^2(\pi t / 2), \quad b(t) = \sin^2(\pi t / 2)
\end{equation}
This scheduling ensures the smoothness and numerical stability of the diffusion process at the start and end points, conducive to training convergence and generation quality.

\subsection{ID Feature Integration Module}
In EditID, we discussed that the integration of ID features significantly affects the model's editability. Specifically, the ID feature integration module is responsible for injecting the mapping features obtained from the character feature extraction module and the shift features generated by the fusion module into the image generation main branch. However, under the train-free framework, this integration often adopts a rigid approach, leading to semantic conflicts, degraded editing capabilities, and unstable identity consistency in complex narrative scenes. To address this challenge, EditIDv2 focuses on optimizing the ID feature integration process, achieving more flexible feature injection by finely decomposing the PerceiverAttention module into multiple controllable sub-stages.

The PerceiverAttention module is essentially a cross-attention mechanism used to bridge the character feature branch and the image generation branch. We first decompose this module into independent computation paths for query (Query), key (Key), and value (Value). The specific formulas are as follows:

\begin{equation}
Q = W_Q \cdot x_{id}, \quad K = W_K \cdot x_{gen}, \quad V = W_V \cdot x_{gen}
\end{equation}

where $x_{id}$ represents the ID feature vector from the character feature branch, $x_{gen}$ represents the current state vector of the image generation branch, and $W_Q, W_K, W_V$ are learnable weights. Through this decomposition, we can independently adjust the strength and depth of each path, for example, introducing an adjustable weight $\alpha$ in the query path to control the influence of ID features:

\begin{equation}
Q' = \alpha \cdot Q + (1 - \alpha) \cdot Q_{noise}
\end{equation}

Here, $Q_{noise}$ is the query compensation term extracted from the noise image, used to inject additional degrees of freedom in the latent space and improve editability. This decomposition allows us to optimize only the attention cross mechanism during the fine-tuning stage without interfering with the overall generalization ability of the DiT architecture. At the same time, we introduce dynamic integration strength control, adaptively adjusting $\alpha$ according to the denoising step $t$:

\begin{equation}
\alpha(t) = \alpha_0 \cdot (1 - t / T)
\end{equation}

where $T$ is the total number of denoising steps, and $\alpha_0$ is the initial strength. This progressive adjustment avoids the problems of image darkening and stability loss caused by blunt strength control in EditID, ensuring that the model can handle multi-level semantic editing, such as actions, expressions, and environmental changes, under long text inputs.

In addition, to support minimal data lubrication, we apply dynamic training on the decomposed PerceiverAttention, using only a small amount of labeled data (for example, image pairs of the same ID with different attributes, such as different poses or expressions) to fine-tune the attention weights. This enables the integration module to gradually inject editability while maintaining ID consistency, forming an efficient "lubrication" process.

\subsection{ID Loss Joint}
To strengthen identity consistency and seamlessly integrate with the denoising process of the diffusion model, we introduce an ID loss function as an auxiliary supervision signal and design a joint dynamic training mechanism for the diffusion model. The ID loss function quantifies identity fidelity by calculating the cosine similarity between the generated image and the reference ID features. Specifically, for the generated latent representation $\hat{x}_0$ and the reference ID embedding $e_{ref}$, the ID loss is defined as:

\begin{equation}
\mathcal{L}_{ID} = 1 - \frac{\hat{x}_0 \cdot e_{ref}}{\|\hat{x}_0\|_2 \cdot \|e_{ref}\|_2}
\end{equation}

This loss function serves as a regularization term in training, ensuring that the integration process does not damage the core features of character identity. At the same time, we jointly optimize the ID loss with the flow matching loss $\mathcal{L}_{diff}$ of the diffusion model, with the total loss being:

\begin{equation}
\mathcal{L}_{total} = \mathcal{L}_{diff} + \lambda \cdot \mathcal{L}_{ID}
\end{equation}

where $\lambda$ is an adjustable weight (set to 0.5 in experiments), used to balance consistency and editability.

The joint dynamic training mechanism dynamically adjusts the ID integration strength and depth in each diffusion denoising step. Specifically, under the flow matching framework, we express the denoising process as the solution path of an ordinary differential equation (ODE) and inject ID supervision at each time step $t$:

\begin{equation}
\frac{dx}{dt} = f_\theta(x_t, t, c) + \beta(t) \cdot g(x_t, e_{ref})
\end{equation}

Here, $f_\theta$ is the original vector field, $g$ is the ID guidance function (based on cosine similarity gradients), and $\beta(t)$ is a dynamic weight using cosine decay scheduling to strengthen consistency in early steps and relax in later steps to enhance editing flexibility. This mechanism solves the semantic conflict problems caused by rigid integration in traditional train-free methods, ensuring that the model can handle temporal logic and contextual relationships in long prompts under high-complexity narrative scenes.

Under minimal data lubrication, this joint training converges with only a small amount of labeled data, injecting editability through progressive iteration while keeping ID loss at a low level. This not only improves the model's robustness but also achieves significant improvements in the editability indicators of IBench.

\subsection{ID Integration Feature Fusion}
To further optimize the efficiency and flexibility of the integration process, we adopt an offline fusion strategy for the integration module. This strategy optimizes only the cross-attention weights of PerceiverAttention during the fine-tuning stage, then fuses multiple optimized weight sets offline to achieve stepwise control of consistency and editability.

Specifically, we first train multiple independent attention weight variants, each targeting different objectives: for example, variant A emphasizes ID consistency (by increasing $\lambda$), and variant B emphasizes editability (by reducing the initial value of $\alpha$). These variants are fused through weighted averaging in the offline stage:

\begin{equation}
W_{fused} = \sum_{i=1}^N w_i \cdot W_i
\end{equation}

where $W_i$ is the weight of the i-th variant, and $w_i$ is the fusion coefficient (automatically adjusted based on validation set performance). This offline fusion avoids the computational overhead of online training and achieves efficient semantic injection without damaging the model's generalization ability.

In addition, the fusion strategy supports dynamic selection: for complex narrative tasks, we prioritize editability weights; for simple ID maintenance tasks, we bias toward consistency weights. This enables EditIDv2 to flexibly adapt to different scenes under long text inputs and activate the fusion module with only minimal data lubrication in actual deployment.

\section{Training Data and Method}
We adopt a tuning-free approach to fine-tune the PerceiverAttention module and propose a data lubrication method, achieving the injection of character ID feature information with only a small amount of data.

\subsection{Training Dataset}
We collected our training data from public datasets such as MyStyle \cite{nitzan2022mystyle}, as well as some publicly crawled data from the internet. They are all images of different attributes under the same ID. For example, in the MyStyle data, we mainly collect images of different facial poses under the same ID. The collected training data is divided into three dimensions: 1. Rich character scene data, including different attributes in character scenes, such as poses, lighting, accessories, etc. Although the number of IDs is small, the number of images under the ID is large, which is the core data requirement for data lubrication; 2. Mainly enhanced data for Asian people in Chinese scenes; 3. Ensure balance of images under the same ID. We used GPT-4o for labeling. After multiple data screenings, we collected a total of about 3K finely labeled training samples.

\begin{figure}[t]
  \centering
  \includegraphics[width=\linewidth]{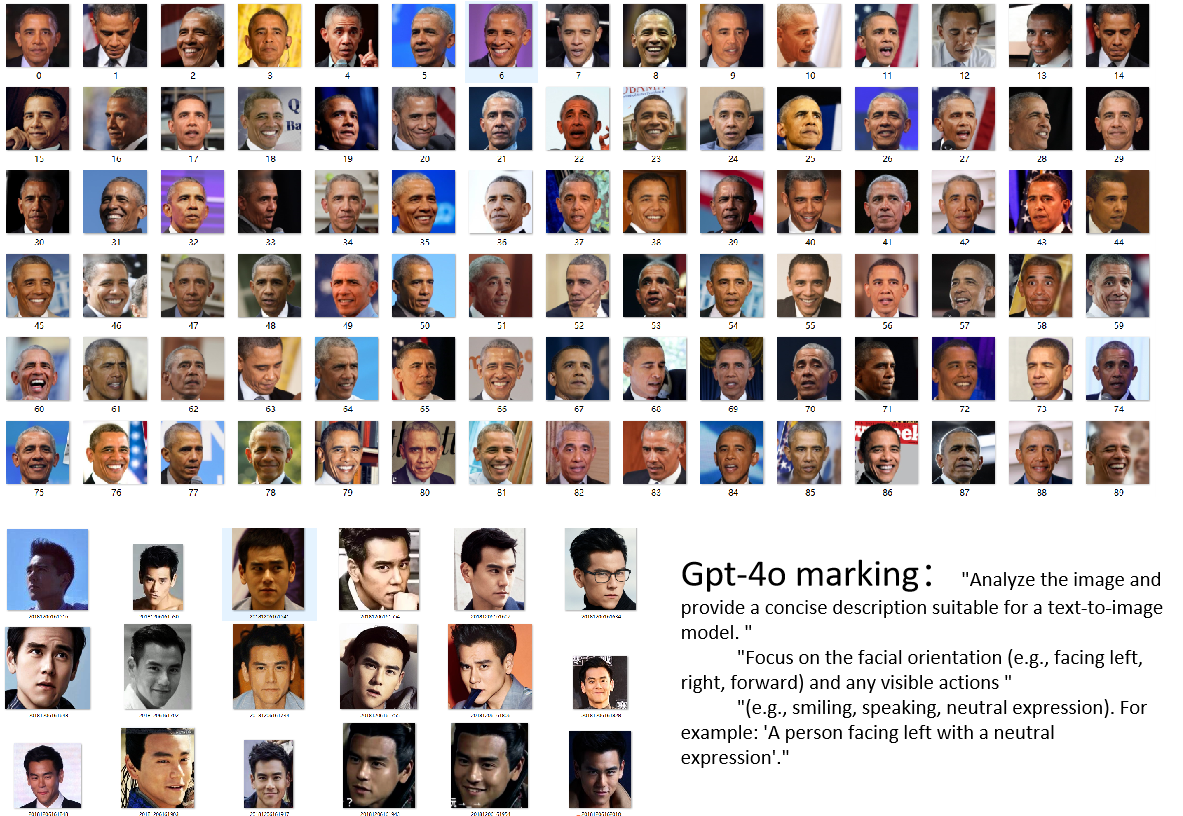}
  \caption{Examples of our training data, where the images are attributes of different poses under the same ID, and the right side is the prompt described by GPT-4o.}
  \label{fig:figure3}
\end{figure}

\subsection{Training Method}
EditIDv2 adopts a tuning-free training strategy, combined with minimal data lubrication (about 3K labeled data, including different poses, lighting, etc., under the same ID), to achieve editability injection by fine-tuning the cross-attention module while ensuring character ID consistency. The training process is as follows: Initialize the Flux model based on the DiT architecture and pre-trained weights, with only the cross-attention module parameters trainable, and the VAE, text encoder, and visual encoder frozen. Input images are encoded into latent representations, text prompts generate conditional embeddings, and ID images extract feature vectors. During training, randomly sample time steps, generate noisy images, predict noise through the model, and balance generation quality and identity consistency through joint optimization of diffusion loss (MSE Loss) and ID loss (cosine similarity loss). The optimizer uses AdamW, and the learning rate is dynamically adjusted through a cosine scheduler. In each iteration, randomly sample time step t, combine flow matching methods to calculate diffusion loss, and extract the cosine similarity between the generated image and the reference ID features through the ID loss module to ensure identity consistency. Offline fusion of multiple optimized weights dynamically balances consistency and editability, effectively supporting high-quality semantic editing in complex narrative scenes.

\section{Experiments}
\label{sec:experiments}

\subsection{Setting}
\label{subsec:setting}

We use the Flux version of PuLID as the base model, which is also the base model for EditID. We did not directly select the shift and mapping features of EditID, but retained the original feature combination of PuLID. For the Flux model, the sampling steps are set to 20, guidance to 3.5, cfg\_scale to 1, and the sampler uses Euler. For the character feature branch, Antelopev2 \cite{deng2019arcface} is used as the face recognition model, and EVA-CLIP \cite{sun2023eva} as the CLIP image encoder. All our experiments are completed on 4 NVIDIA A100 GPUs, with the inference framework being ComfyUI. Our evaluation adopts the IBench framework proposed by EditID.

\subsection{Qualitative Comparison}
\label{subsec:qualitative_comparison}

We adopt the SDXL version of PuLID, the Flux version of PuLID, InfiniteYou \cite{jiang2025infiniteyou}, and EditID as the comparative model group, where the base model for PuLID SDXL is sdxl\_base\_1.0 \cite{podell2023sdxl}. The base models for InfiniteYou and EditID are both Flux.1 dev. As shown in Figure 4, EditIDv2 takes long text prompts as input and, while maintaining character consistency, achieves better editability compared to the almost synchronous paste-copy input face of PuLID Flux; compared to EditID, it not only maintains quite good editability, but the generated images are more natural and follow the prompt instructions. The following four groups of prompts are modern urban suspense, mythological epic battlefield, rural pastoral memories, and steampunk adventure, four complex narrative scenes. In the first column, both EditID and EditIDv2 show good descriptive abilities. In the second column, when "Her braided hair sways in the wind, and her fierce eyes scan the clashing armies below." this scene description, EditIDv2 shows good editability, achieving the action of overlooking, while EditID and InfiniteYou as well as PuLID fail to accurately express the concepts in the narrative scene. In the fourth column, EditID loses the concept of goggles, InfiniteYou retains goggles but is far from EditIDv2's performance of "and her face shows a mix of excitement and concentration as she navigates through swirling clouds". It can be seen that EditIDv2 accurately grasps semantics and fully presents editability in such complex narrative scenes. The ID generation quality and detail handling of the PuLID SDXL version are generally poor, with low fidelity, possibly due to the limited generation capability of the SDXL base model. The character consistency is also inferior to the Flux version, and the generated images have a weak non-realistic scene style attribute.

\begin{figure*}[t]
    \centering
    \includegraphics[width=\textwidth]{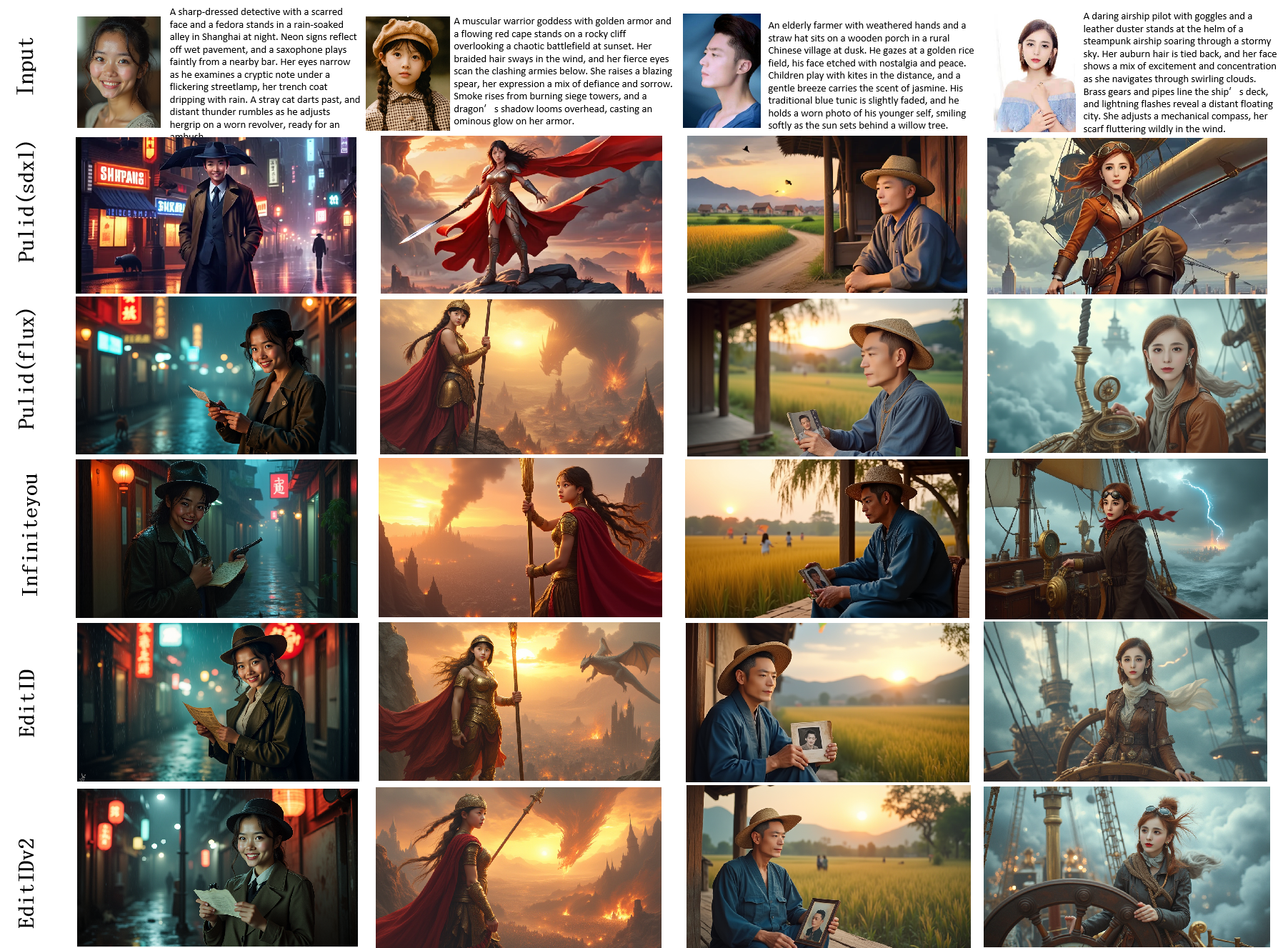}
    \caption{Qualitative Comparison: EditIDv2 achieves higher editability while ensuring ID consistency. EditIDv2 can accurately implement concept expression in complex narrative scenes and fully present editability.}
    \label{fig:qualitative_comparison}
\end{figure*}

We mainly focus on the performance of EditIDv2 on long text prompts. In the table, using the evaluation combination of ChineseID + editable long prompts, EditIDv2 reaches 0.691 on the Aesthetic metric, surpassing all comparative models, demonstrating higher aesthetic quality in complex narrative scenes. At the same time, although Image quality is 0.437, this reflects its priority optimization for editability rather than simply pursuing static beauty. In Facesim, which measures consistency, EditIDv2 is 0.659, a decrease of about 0.076 compared to the Flux version of PuLID, but according to Figure 8 in the paper and related analysis, the Flux version of PuLID has overly strong consistency, even producing a paste-copy phenomenon of the output ID face and input ID face, which greatly limits the applicability of text-image consistency generation. Facial and limb features need to produce different changes in different scenes, and EditIDv2 cleverly balances this problem through the data lubrication mechanism, avoiding excessive constraints. In ClipI, EditIDv2 reaches 0.804, indicating minimal interference of ID insertion on the original generation capability, even higher than the EditID and InfiniteYou series, demonstrating its efficient feature integration strategy. On ClipT text following ability, it is 0.253, also maintaining a good level. On the most important editability indicators, the three Euler angles of Posediv in EditIDv2 produce a Yaw improvement of about 8.872, Pitch improvement of about 3.769, Roll improvement of about 1.917 compared to the Flux version of PuLID, totaling over 14 points improvement. Landmarkdiff reaches 0.096, significantly higher than PuLID Flux's 0.070, and Exprdiv is 0.611, superior to most models, indicating that EditIDv2 sacrifices only a little similarity (limited decrease in Facesim) but brings excellent editability. In fact, the character consistency constraint in PuLID Flux is too strong, and releasing excessive consistency in exchange for improved editability is an extremely wise choice, especially under high-complexity narratives and long text inputs, which enables EditIDv2 to better handle multi-level semantic editing, such as poses, expressions, and environmental changes. Compared to the SDXL version of PuLID, although editability is high, character consistency is sacrificed too much (Facesim only 0.399), while EditIDv2 pushes Posediv and Landmarkdiff to new highs while maintaining high consistency; compared to the InfiniteYou series, although its Posediv is higher, Fid is too high (40+), indicating unstable generation quality, and ClipI and ClipT are not as good as EditIDv2, proving that EditIDv2's PerceiverAttention decomposition and offline fusion strategy more effectively inject editability under minimal data lubrication, achieving SOTA results in IBench.

\begin{table*}[t]
    \centering
    \caption{Evaluation metric results from IBench on ChineseID with editable long prompts}
    \label{tab:chineseid_evaluation}
    \begin{tabular}{lccccccccc}
        \toprule
        Model & FID & Aesthetic & Image Quality & \multicolumn{3}{c}{Posediv} & Landmarkdiff & Exprdiv \\
        \cmidrule(lr){5-7}
        & & & & Yaw & Pitch & Roll & & \\
        \midrule
        PuLID (SDXL) & 11.28 & 0.675 & 0.502 & 19.48 & 6.187 & 12.69 & 0.099 & 0.593 \\
        PuLID (Flux) & 14.59 & 0.681 & 0.431 & 9.298 & 5.872 & 9.473 & 0.070 & 0.562 \\
        EditID & 13.52 & 0.683 & 0.454 & 11.81 & 6.722 & 10.60 & 0.082 & 0.554 \\
        InfiniteYou (aes) & 40.70 & 0.676 & 0.523 & 24.85 & 9.322 & 12.34 & 0.083 & 0.616 \\
        InfiniteYou (sim) & 43.157 & 0.675 & 0.522 & 25.02 & 9.431 & 12.30 & 0.082 & 0.635 \\
        EditIDv2 & - & 0.691 & 0.437 & 18.17 & 9.641 & 11.39 & 0.096 & 0.611 \\
        \midrule
        & Facesim & ClipI & ClipT & Dino & Fgis & & & \\
        \cmidrule(lr){2-6}
        PuLID (SDXL) & 0.399 & 0.768 & 0.248 & 0.129 & 0.353 & & & \\
        PuLID (Flux) & 0.735 & 0.757 & 0.243 & 0.178 & 0.501 & & & \\
        EditID & 0.714 & 0.769 & 0.249 & 0.162 & 0.459 & & & \\
        InfiniteYou (aes) & 0.547 & 0.794 & 0.276 & - & - & & & \\
        InfiniteYou (sim) & 0.552 & 0.808 & 0.257 & - & - & & & \\
        EditIDv2 & 0.659 & 0.804 & 0.253 & - & - & & & \\
        \bottomrule
    \end{tabular}
\end{table*}

\subsection{Ablation Study}
\subsubsection{Comparison of Loss Function Designs}
To verify the impact of the ID loss function on model performance, we compared training schemes without ID loss and with joint ID loss and diffusion loss. Without ID loss, the model's generation results in complex narrative scenes show strong semantic biases, especially in long text prompts, where the identity consistency of generated images is weak, and character features (such as facial details, expressions) do not match the reference ID sufficiently, with limited editability, making it difficult to accurately reflect actions or environmental changes in the prompts. In contrast, the joint training scheme of ID loss and diffusion loss significantly improves generation quality. By introducing ID loss (supervision signal based on cosine similarity) in the denoising process, the model can enhance semantic understanding capabilities in complex narrative scenes while maintaining character identity consistency. For example, in Figure 5, the images generated by joint training are more consistent with the input prompts in details of poses, expressions, and backgrounds, showing more natural transitions and higher editability, while the generation results without ID loss appear relatively rigid and lack semantic layering.

\begin{figure}[t]
\centering
\includegraphics[width=\linewidth]{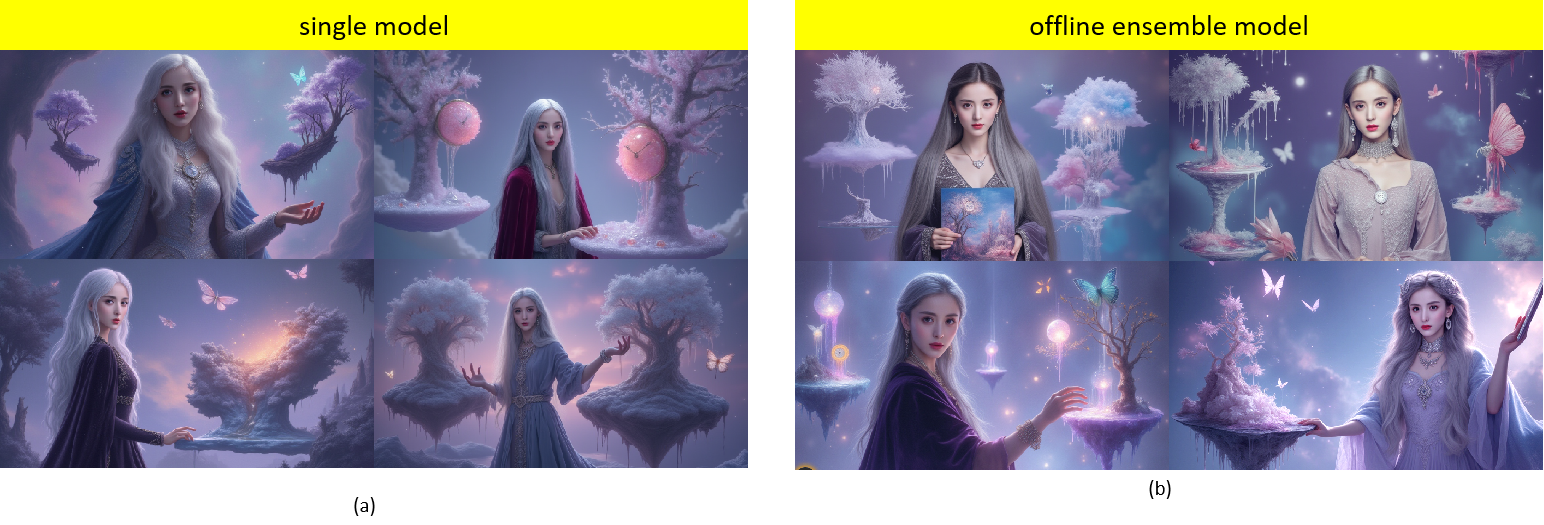}
\caption{Comparison of results with ID loss and joint training of ID loss and diffusion loss.}
\label{fig:figure5}
\end{figure}

\subsubsection{Comparison of Model Offline Integration Strategies}
We further compared the effects of single model weights and multi-model offline fusion strategies. Single model weights have difficulty achieving a balance between consistency and editability when processing long text inputs, often showing insufficient semantic response to prompts, especially in complex narrative scenes (such as the dream surreal scene in the figure below), where the generated actions or expression details are not rich enough. The multi-model offline fusion strategy significantly improves the model's flexibility by fusing multiple optimized attention weights. As shown in Figures 4 and 6, the images generated by multi-model fusion better present the multi-level semantics in the prompts while maintaining character identity consistency, for example, in "overlooking the battlefield" or "steampunk adventure" scenes, the dynamic poses and environmental atmospheres of characters are better presented, with the overall picture having more narrative sense and immersion, proving the superiority of the offline fusion strategy in complex scenes.

\begin{figure}[t]
\centering
\includegraphics[width=\linewidth]{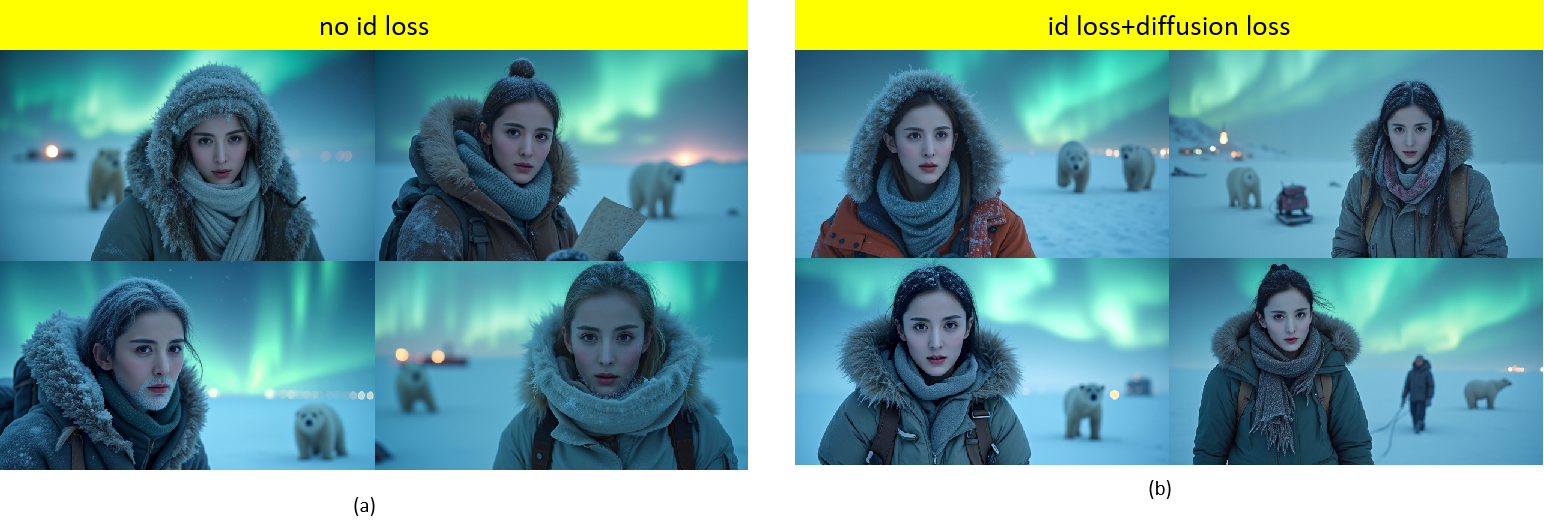}
\caption{Comparison of one single model of EditIDv2 and the results after offline integration of multiple weights.}
\label{fig:figure6}
\end{figure}

\section{Conclusion}
This paper proposes EditIDv2, a tuning-free ID customization method for text-to-image generation. We further explore injecting editability through minimal data lubrication in the DiT architecture and achieve state-of-the-art performance under high-complexity narrative scenes and long prompt inputs. Based on the Flux model, we focus on optimizing the ID feature embedding module, solving semantic conflict problems caused by rigid embedding in traditional train-free methods through a fine decomposition of PerceiverAttention, the introduction of an ID loss function, joint dynamic training of the diffusion model, and an offline fusion strategy, thereby significantly improving the editability of character customization. Our method requires only a small amount of labeled data as lubrication to balance identity consistency and editing flexibility, demonstrating its potential for efficiently generating high-quality personalized images in complex contextual environments. Moreover, this framework can be extended to other ID customization algorithms equipped with character feature branches. In future work, we will continue to investigate the application of data lubrication mechanisms on larger-scale datasets and design more advanced dynamic embedding strategies to further synchronously optimize the performance of consistency and editability in long-sequence narratives. We believe that through sophisticated embedding optimization, EditIDv2 will provide broader application prospects for immersive content creation and personalized generation.

{\small
\bibliographystyle{ieee_fullname}
\bibliography{egbib}

\begin{thebibliography}{10}\itemsep=-1pt

\bibitem{agro2024uno}
Ben Agro, Quinlan Sykora, Sergio Casas, Thomas Gilles, and Raquel Urtasun.
\newblock Uno: Unsupervised occupancy fields for perception and forecasting.
\newblock In {\em Proceedings of the IEEE/CVF Conference on Computer Vision and Pattern Recognition}, pages 14487--14496, 2024.

\bibitem{batifol2025flux}
Stephen Batifol, Andreas Blattmann, Frederic Boesel, Saksham Consul, Cyril Diagne, Tim Dockhorn, Jack English, Zion English, Patrick Esser, Sumith Kulal, et~al.
\newblock Flux. 1 kontext: Flow matching for in-context image generation and editing in latent space.
\newblock {\em arXiv e-prints}, pages arXiv--2506, 2025.

\bibitem{chen2025xverse}
Bowen Chen, Mengyi Zhao, Haomiao Sun, Li Chen, Xu Wang, Kang Du, and Xinglong Wu.
\newblock Xverse: Consistent multi-subject control of identity and semantic attributes via dit modulation.
\newblock {\em arXiv preprint arXiv:2506.21416}, 2025.

\bibitem{chen2025unireal}
Xi Chen, Zhifei Zhang, He Zhang, Yuqian Zhou, Soo~Ye Kim, Qing Liu, Yijun Li, Jianming Zhang, Nanxuan Zhao, Yilin Wang, et~al.
\newblock Unireal: Universal image generation and editing via learning real-world dynamics.
\newblock In {\em Proceedings of the Computer Vision and Pattern Recognition Conference}, pages 12501--12511, 2025.

\bibitem{deng2019arcface}
Jiankang Deng, Jia Guo, Niannan Xue, and Stefanos Zafeiriou.
\newblock Arcface: Additive angular margin loss for deep face recognition.
\newblock In {\em Proceedings of the IEEE/CVF conference on computer vision and pattern recognition}, pages 4690--4699, 2019.

\bibitem{esser2024scaling}
Patrick Esser, Sumith Kulal, Andreas Blattmann, Rahim Entezari, Jonas M{\"u}ller, Harry Saini, Yam Levi, Dominik Lorenz, Axel Sauer, Frederic Boesel, et~al.
\newblock Scaling rectified flow transformers for high-resolution image synthesis.
\newblock In {\em Forty-first international conference on machine learning}, 2024.

\bibitem{gal2022image}
Rinon Gal, Yuval Alaluf, Yuval Atzmon, Or Patashnik, Amit~H Bermano, Gal Chechik, and Daniel Cohen-Or.
\newblock An image is worth one word: Personalizing text-to-image generation using textual inversion.
\newblock {\em arXiv preprint arXiv:2208.01618}, 2022.

\bibitem{guo2024pulid}
Zinan Guo, Yanze Wu, Chen Zhuowei, Peng Zhang, Qian He, et~al.
\newblock Pulid: Pure and lightning id customization via contrastive alignment.
\newblock {\em Advances in neural information processing systems}, 37:36777--36804, 2024.

\bibitem{jiang2025infiniteyou}
Liming Jiang, Qing Yan, Yumin Jia, Zichuan Liu, Hao Kang, and Xin Lu.
\newblock Infiniteyou: Flexible photo recrafting while preserving your identity.
\newblock {\em arXiv preprint arXiv:2503.16418}, 2025.

\bibitem{kumari2023multi}
Nupur Kumari, Bingliang Zhang, Richard Zhang, Eli Shechtman, and Jun-Yan Zhu.
\newblock Multi-concept customization of text-to-image diffusion.
\newblock In {\em Proceedings of the IEEE/CVF conference on computer vision and pattern recognition}, pages 1931--1941, 2023.

\bibitem{li2025editid}
Guandong Li and Zhaobin Chu.
\newblock Editid: Training-free editable id customization for text-to-image generation.
\newblock {\em arXiv preprint arXiv:2503.12526}, 2025.

\bibitem{lipman2022flow}
Yaron Lipman, Ricky~TQ Chen, Heli Ben-Hamu, Maximilian Nickel, and Matt Le.
\newblock Flow matching for generative modeling.
\newblock {\em arXiv preprint arXiv:2210.02747}, 2022.

\bibitem{liu2023facechain}
Yang Liu, Cheng Yu, Lei Shang, Yongyi He, Ziheng Wu, Xingjun Wang, Chao Xu, Haoyu Xie, Weida Wang, Yuze Zhao, et~al.
\newblock Facechain: A playground for human-centric artificial intelligence generated content.
\newblock {\em arXiv preprint arXiv:2308.14256}, 2023.

\bibitem{mou2025dreamo}
Chong Mou, Yanze Wu, Wenxu Wu, Zinan Guo, Pengze Zhang, Yufeng Cheng, Yiming Luo, Fei Ding, Shiwen Zhang, Xinghui Li, et~al.
\newblock Dreamo: A unified framework for image customization.
\newblock {\em arXiv preprint arXiv:2504.16915}, 2025.

\bibitem{nitzan2022mystyle}
Yotam Nitzan, Kfir Aberman, Qiurui He, Orly Liba, Michal Yarom, Yossi Gandelsman, Inbar Mosseri, Yael Pritch, and Daniel Cohen-Or.
\newblock Mystyle: A personalized generative prior.
\newblock {\em ACM Transactions on Graphics (TOG)}, 41(6):1--10, 2022.

\bibitem{peebles2023scalable}
William Peebles and Saining Xie.
\newblock Scalable diffusion models with transformers.
\newblock In {\em Proceedings of the IEEE/CVF international conference on computer vision}, pages 4195--4205, 2023.

\bibitem{peng2024portraitbooth}
Xu Peng, Junwei Zhu, Boyuan Jiang, Ying Tai, Donghao Luo, Jiangning Zhang, Wei Lin, Taisong Jin, Chengjie Wang, and Rongrong Ji.
\newblock Portraitbooth: A versatile portrait model for fast identity-preserved personalization.
\newblock In {\em Proceedings of the IEEE/CVF Conference on Computer Vision and Pattern Recognition}, pages 27080--27090, 2024.

\bibitem{podell2023sdxl}
Dustin Podell, Zion English, Kyle Lacey, Andreas Blattmann, Tim Dockhorn, Jonas M{\"u}ller, Joe Penna, and Robin Rombach.
\newblock Sdxl: Improving latent diffusion models for high-resolution image synthesis.
\newblock {\em arXiv preprint arXiv:2307.01952}, 2023.

\bibitem{ruiz2023dreambooth}
Nataniel Ruiz, Yuanzhen Li, Varun Jampani, Yael Pritch, Michael Rubinstein, and Kfir Aberman.
\newblock Dreambooth: Fine tuning text-to-image diffusion models for subject-driven generation.
\newblock In {\em Proceedings of the IEEE/CVF conference on computer vision and pattern recognition}, pages 22500--22510, 2023.

\bibitem{sun2023eva}
Quan Sun, Yuxin Fang, Ledell Wu, Xinlong Wang, and Yue Cao.
\newblock Eva-clip: Improved training techniques for clip at scale.
\newblock {\em arXiv preprint arXiv:2303.15389}, 2023.

\bibitem{tan2024ominicontrol}
Zhenxiong Tan, Songhua Liu, Xingyi Yang, Qiaochu Xue, and Xinchao Wang.
\newblock Ominicontrol: Minimal and universal control for diffusion transformer.
\newblock {\em arXiv preprint arXiv:2411.15098}, 2024.

\bibitem{tewel2024training}
Yoad Tewel, Omri Kaduri, Rinon Gal, Yoni Kasten, Lior Wolf, Gal Chechik, and Yuval Atzmon.
\newblock Training-free consistent text-to-image generation.
\newblock {\em ACM Transactions on Graphics (TOG)}, 43(4):1--18, 2024.

\bibitem{wu2025uso}
Shaojin Wu, Mengqi Huang, Yufeng Cheng, Wenxu Wu, Jiahe Tian, Yiming Luo, Fei Ding, and Qian He.
\newblock Uso: Unified style and subject-driven generation via disentangled and reward learning.
\newblock {\em arXiv preprint arXiv:2508.18966}, 2025.

\bibitem{wu2025less}
Shaojin Wu, Mengqi Huang, Wenxu Wu, Yufeng Cheng, Fei Ding, and Qian He.
\newblock Less-to-more generalization: Unlocking more controllability by in-context generation.
\newblock {\em arXiv preprint arXiv:2504.02160}, 2025.

\bibitem{xiao2025fastcomposer}
Guangxuan Xiao, Tianwei Yin, William~T Freeman, Fr{\'e}do Durand, and Song Han.
\newblock Fastcomposer: Tuning-free multi-subject image generation with localized attention.
\newblock {\em International Journal of Computer Vision}, 133(3):1175--1194, 2025.

\bibitem{ye2023ip}
Hu Ye, Jun Zhang, Sibo Liu, Xiao Han, and Wei Yang.
\newblock Ip-adapter: Text compatible image prompt adapter for text-to-image diffusion models.
\newblock {\em arXiv preprint arXiv:2308.06721}, 2023.

\end{thebibliography}
}

\end{document}